%% file: main.tex
\documentclass[letterpaper, 10 pt, conference]{ieeeconf}  %

\IEEEoverridecommandlockouts                              %

\input{0-defs}
\input{0-preamble}

\input{corl_2024/tables/tables}
\input{corl_2024/figures_tex/main_figures}

\title{\LARGE \bf
In-Context Imitation Learning via Next-Token Prediction
}

\author{Max (Letian) Fu$^{1}$*, Huang Huang$^{1}$*, Gaurav Datta$^{1}$*, Lawrence Yunliang Chen$^{1}$ \\Will Panitch$^{1}$, Fangchen Liu$^{1}$, Hui Li$^{2}$, Ken Goldberg$^{1}$
\thanks{*Equal contribution, $^{1}$University of California, Berkeley, $^{2}$Autodesk Research}%
}

\begin{document}

\makeatletter
\g@addto@macro\@maketitle{
  \def\mycolspace{0.8\textwidth}
  \centering
 	\begin{tabular}{c@{\hspace{0.01\textwidth}} c@{\hspace{\mycolspace}}}
 	  \centering
 	  &
\includegraphics[width=0.98\linewidth]{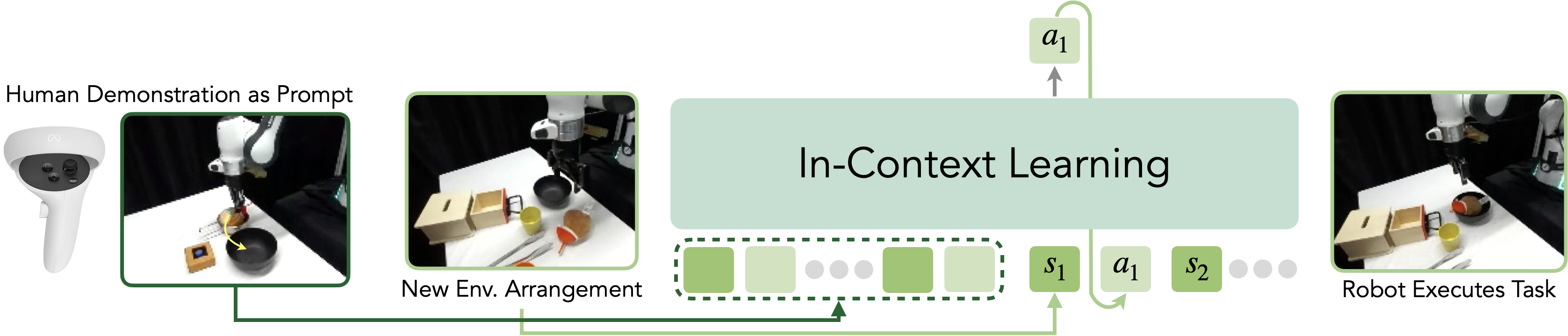}\\
	\end{tabular}
	\captionof{figure}{\textbf{\algname (\algabbr)}: A robot foundation model with in-context imitation learning capabilities. ICRT performs next-token prediction on large-scale sensorimotor trajectories. At inference time, it takes raw sensorimotor trajectories of human teleoperation demonstrations as prompts, enabling the model to execute new tasks with real-time continuous control, without requiring fine-tuning.
    }
\label{fig:splash}
\vspace{-1mm}
}
\makeatother
\maketitle

\thispagestyle{empty}
\pagestyle{empty}

\setcounter{figure}{1}      
\input{sections/0-abstract}

\input{sections/1-intro}

\input{sections/2-related-works}
\input{sections/3-problem-statement}

\input{sections/4-method}

\input{sections/5-experiments}

\input{sections/6-conclusion}
\input{sections/8-acknowledgement}

\renewcommand*{\bibfont}{\footnotesize}
\printbibliography
\clearpage
\input{sections/7-supplement}

\end{document}

%% file: 0-preamble.tex
\usepackage{amsmath}

\usepackage[table]{xcolor}

\usepackage{microtype}
\usepackage{graphicx}
\usepackage{booktabs} %

\usepackage{hyperref}

\usepackage{microtype}
\usepackage{graphicx}
\usepackage{booktabs} %

\usepackage{mathtools}
\usepackage{algorithm}
\usepackage{tikz}
\usetikzlibrary{arrows,automata,positioning}
\usepackage{multirow} 
\usepackage{xspace}

\usepackage{comment}
\usepackage{amssymb} %

\usepackage{dsfont}
\usepackage{tabulary}
\usepackage{tabularx}
\usepackage{makecell}
\usepackage{wrapfig}

\usepackage{dsfont}
\usepackage{tabulary}
\usepackage{makecell}
\usepackage{wrapfig}
\usepackage{subcaption}

\usepackage{amssymb}
\usepackage{mathtools}
\usepackage{amsthm}
\usepackage{array}

\usepackage[capitalize,noabbrev]{cleveref}

\theoremstyle{plain}
\newtheorem{theorem}{Theorem}[section]

\theoremstyle{definition}

\theoremstyle{remark}
\newtheorem{remark}[theorem]{Remark}

\usepackage[textsize=tiny]{todonotes}

\newcommand{\reb}[1]{{\color{black}{#1}}}

\newcommand{\algname}{In-Context Robot Transformer\xspace}
\newcommand{\algabbr}{ICRT\xspace}

\definecolor{orange}{rgb}{1,0.5,0}
\definecolor{lightsalmonpink}{rgb}{1.0, 0.6, 0.6}
\definecolor{verylightsalmonpink}{rgb}{0.966, 0.805, 0.797}
\definecolor{lightblue}{rgb}{0.862, 0.906, 0.984}
\definecolor{lightyellow}{rgb}{1.0, 0.945, 0.797}
\definecolor{lightgreen}{rgb}{0.835, 0.91, 0.828}
\definecolor{lightpurple}{rgb}{0.879, 0.832, 0.902}

\newcommand{\fsize}{small}
\usepackage[font=\fsize,labelfont=bf,skip=5pt]{caption}
\usepackage[export]{adjustbox}
\usepackage{wrapfig}

\usepackage{ragged2e}

\usepackage{listings}

\definecolor{codegreen}{rgb}{0,0.6,0}
\definecolor{codegray}{rgb}{0.5,0.5,0.5}
\definecolor{codepurple}{rgb}{0.58,0,0.82}
\definecolor{backcolour}{rgb}{0.95,0.95,0.92}

\lstdefinestyle{mystyle}{
    backgroundcolor=\color{backcolour},   
    commentstyle=\color{codegreen},
    keywordstyle=\color{magenta},
    numberstyle=\tiny\color{codegray},
    stringstyle=\color{codepurple},
    basicstyle=\ttfamily\footnotesize,
    breakatwhitespace=false,         
    breaklines=true,                 
    captionpos=b,                    
    keepspaces=true,                 
    numbers=left,                    
    numbersep=5pt,                  
    showspaces=false,                
    showstringspaces=false,
    showtabs=false,                  
    tabsize=2
}

\usepackage{url}

\usepackage{hyperref}
\hypersetup{
    colorlinks=true,
    linkcolor=black,
    citecolor=black,
    filecolor=cyan,
    urlcolor=black
}

\usepackage[backend=biber,
            url=false,
            isbn=false,
            doi=false,
            backref=false,
            style=ieee,
            natbib=true,%
            mincitenames=1,
            maxcitenames=1,
            citestyle=numeric-comp,
            sorting=none,%
            block=none]{biblatex}
\renewcommand{\bibfont}{\small}
\renewcommand{\baselinestretch}{1.0}



\newcolumntype{C}[1]{>{\centering\arraybackslash}p{#1}}

%% file: corl_2024/tables/tables.tex
\def\tabResultsTable#1{
    \begin{table}[#1]
    \centering
    \begin{tabular}{lccc}
    \hline
                   & Pick and Place           & Poke                     & Average                  \\ \hline
    Goal Condition & 33.3 ($\pm$6.5)          & 6.7 ($\pm$4.6)           & 20.0 ($\pm$4.3)          \\
    Octo~\cite{octo_2023}           & 5.0 ($\pm$2.7)           & 13.3 ($\pm$6.2)          & 9.2 ($\pm$3.5)           \\
    OpenVLA~\cite{kim2024openvlaopensourcevisionlanguageactionmodel}        & 11.7 ($\pm$4.6)          & 3.3 ($\pm$3.3)           & 7.5 ($\pm$2.9)           \\ \hline
    ICRT           & \textbf{65.0 ($\pm$7.3)} & \textbf{93.3 ($\pm$4.6)} & \textbf{79.2 ($\pm$4.6)} \\ \hline
    \end{tabular}
    \caption{\textbf{Main Results.} \algabbr outperforms two state-of-the-art robot foundation models that are conditioned on goals or language in both pick-and-place and poking tasks. We evaluated each task primitive using six tasks not seen during training, conducting five trials per task for a total of 30 trials per primitive and 60 trials overall to calculate average performance. For each model, we report the mean success rate for each task, the overall success rate, and the corresponding standard error in parentheses.}
    \label{tab:results_table}
    \vspace{1mm}
    \end{table}
}

\def\tabAblationTable#1{
    \begin{table}[#1]
    \begin{tabular}{lccc}
    \hline
                 & Pick and Place           & Poke                     & Average                  \\ \hline
    ICRT-Llama2  & 43.3 ($\pm$7.9)          & 73.3 ($\pm$8.2)          & 58.3 ($\pm$6.0)          \\
    ICRT (DROID) & 0.0 ($\pm$0.0)               & 0.0 ($\pm$0.0)               & 0.0 ($\pm$0.0)               \\
    ICRT (MT)    & \textbf{76.7 ($\pm$7.1)} & 70.0 ($\pm$8.5)          & 73.3 ($\pm$5.5)          \\
    ICRT +Prompt Loss & \multicolumn{1}{l}{21.7 ($\pm$6.2)} & 23.3 ($\pm$7.9) & 22.5 ($\pm$5.0) \\ \hline
    ICRT         & 65.0 ($\pm$7.3)          & \textbf{93.3 ($\pm$4.6)} & \textbf{79.2 ($\pm$4.6)} \\ \hline
    \end{tabular}
    \caption{\textbf{Ablation Study.} We ablate three key design choices: fine-tuning \algabbr using a language model, training solely on the relevant dataset or the fine-tuning dataset, and the impact of including prompt loss in the training process. We report the mean success rates and their corresponding standard errors for the pick-and-place and poke tasks, as well as the overall average performance.}
    \label{tab:ablation_table}
    \vspace{1mm}
    \end{table}
}

\def\tabPickPlace#1{
\begin{table*}[]
\centering
\small
\begin{adjustbox}{max width=\textwidth}

\begin{tabular}{lcccccc|c}
\hline
\begin{tabular}[c]{@{}l@{}}Pick Object\\ Place Location\end{tabular} & \begin{tabular}[c]{@{}c@{}}Yellow Cube\\ Black Bowl\end{tabular} & \begin{tabular}[c]{@{}c@{}}Yellow Cube\\ Grey Bowl\end{tabular} & \begin{tabular}[c]{@{}c@{}}Blue Bear\\ Pink Bowl\end{tabular} & \begin{tabular}[c]{@{}c@{}}Radish\\ Grey Bowl\end{tabular} & 
\begin{tabular}[c]{@{}c@{}}Black Dog\\ Pink Bowl\end{tabular} & \begin{tabular}[c]{@{}c@{}}Blue Sponge\\ Silver Pot\end{tabular} & Average Success ($\pm$ Std Err.)\\ 
\hline
Goal Conditioned    & 40\%     & 30\%  & 20\%      & 40\%         & 40\%         & 30\%  &  \reb{33.3\% ($\pm$6.5\%) }  \\ 

\reb{Octo}   & \reb{10\%} & \reb{0\%} & \reb{10\%} &\reb{ 10\%  }        & \reb{0\%     }      & \reb{0\%  }  &    \reb{5.0\% ($\pm$2.7\%) } \\
\reb{OpenVLA } &\reb{ 0\% }& \reb{0\% }& \reb{0\%  }        & \reb{50\%  }        & \reb{20\% }          & \reb{0\%  }  &    \reb{11.7\% ($\pm$4.6\%)  } \\

\hline
\algabbr-Llama2  & 40\%       & 40\%     & 40\%    & 60\%     & 40\%     & 40\%  & \reb{43.3\% ($\pm$7.9\%)}\\
\algabbr (DROID)  & 0\%       & 0\%     & 0\%    & 0\%     & 0\%     & 0\%  & \reb{0.0\% ($\pm$0.0\%)}\\
\algabbr (MT)    & \textbf{90\%}   & \textbf{50\%}   & \textbf{80\%}        & \textbf{90\%}   & \textbf{60\%}     & \textbf{90\%}   & \reb{\textbf{76.7\% ($\pm$7.1\%)} }\\ 
\algabbr+Prompt Loss                                    & 20\%                                                    & 10\%                                                   & 20\%                                                 & 40\%                                              & 30\%                                                 & 10\%                                                  & 21.7\% ($\pm$6.2\%) \\
\algabbr (Co-train)                                    & 10\%                                                    & 0\%                                                   & 10\%                                                 & 0\%                                              & 40\%                                                 & 20\%                                                 &  13.3\% ($\pm$5.8\%) \\
\hline
\algabbr   & 60\%   & \textbf{50\%}      & \textbf{80\%}     & 50\%  & \textbf{60\%}     & \textbf{90\%}   & \reb{65.0\% ($\pm$7.3\%)} \\
\hline
\end{tabular}

 \end{adjustbox}
\caption{Experimental results for the \textit{pick-and-place} primitive. Here we list the 6 tasks that were evaluated and their corresponding success rate. For each model, we also report the mean success rate and the standard error.}
\label{tab:pick_place}
\end{table*}
}

\def\tabPoke#1{
\begin{table*}[#1]
\centering
\small
\begin{adjustbox}{max width=\textwidth}
\begin{tabular}{lcccccc|c}
\hline
Poke Object                       & Radish         & Red Cube       & Grey Dog      & Black Cube    & Pink Bowl      & Blue Sponge    & Average Success ($\pm$ Std Err.)\\ 
\hline
Goal Conditioned                  & 0\%            & 0\%            & 0\%           & 0\%           & 40\%           & 0\%     &   \reb{ 6.7\% ($\pm$4.6\%)}   \\

\reb{Octo}   & \reb{20\%} & \reb{0\%} & \reb{60\%} &\reb{ 0\%  }        & \reb{0\%     }      & \reb{0\%  }  &    \reb{13.3\% ($\pm$6.2\%) } \\
\reb{OpenVLA} &\reb{ 20\% }& \reb{0\% }& \reb{0\%  }        & \reb{0\%  }        & \reb{0\% }          & \reb{0\%  }  &    \reb{3.3\% ($\pm$3.3\%)  } \\

\hline
\algabbr-Llama2    & 60\%           & \textbf{100\%} & 80\%          & 60\%          & 60\%           & 80\%       &   \reb{73.3\% ($\pm$8.2\%)} \\
\algabbr (DROID)  & 0\%       & 0\%     & 0\%    & 0\%     & 0\%     & 0\%  & \reb{0.0\% ($\pm$0.0\%)}\\
\algabbr (MT) & \textbf{100\%} & \textbf{100\%} & 40\%          & 60\%          & 60\%           & 60\%       &   \reb{70.0\% ($\pm$8.5\%) }\\ 
\algabbr+Prompt Loss & 0\% & 20\% & 20\%          & \textbf{80\%}          & 0\%           & 20\%           & 23.3\% ($\pm$7.9\%)\\
\algabbr (Co-train) & 0\% & 0\% & 0\%          & 0\%          & 0\%           & 0\%           & 0.0\% ($\pm$0.0\%)\\
\hline
\algabbr           & \textbf{100\%} & \textbf{100\%} & \textbf{80\%} & \textbf{80\%} & \textbf{100\%} & \textbf{100\%} & \reb{\textbf{93.3\% ($\pm$4.6\%) }}\\ \hline
\end{tabular}
 \end{adjustbox}
\caption{Experimental results for the \textit{poking} primitive. Here we list the 6 tasks that were evaluated and their corresponding success rate. For each model, we also report the mean success rate and the standard error.}
\label{tab:poke}
\end{table*}

}

\def\tabStd#1{
\begin{table}[#1]
\centering
\normalsize
\begin{tabular}{lcc}
\toprule 
Task & Pick and Place Block Dog in Pink Bowl & Poke Blue Sponge \\
\hline 
Success Rate Ave. $\pm$ Std.& 60\% $\pm$ 0.5\% & 88\% $\pm$ 3.2\% \\
\bottomrule
\end{tabular}
\caption{\reb{Repeatability experiments for a pick and place task and a poking task. Each task is conducted by 5 rollouts and each rollout contains 5 trials, resulting in a total of 25 trials.}}
\label{tab:std}
\end{table}
}

\def\tabPrompt#1{
\begin{table}[#1]
\centering
\normalsize
\begin{adjustbox}{max width=\textwidth}
\begin{tabular}{l|ccc|cc}
\toprule 
Prompt Type & No Distractor & One Distractor & Distractor Placement & Two Prompts & Three Prompts\\
\hline
Success Rate & 60\%  & 80\% & 70\% & 80\% & 80\% \\
\bottomrule
\end{tabular}
 \end{adjustbox}
\caption{\reb{Experiments on different prompt types on a pick up black dog and place in the pink bowl task. The first three columns are results for a single prompt trajectory of different types, while the last two columns are that for using two and three prompts. Success rates are calculated over 5 trials for each experiment.}}
\label{tab:prompt}
\vspace{-0.6cm}
\end{table}
}

\def\tabPrimitive#1{
\begin{table}[#1]
\centering
\normalsize
\begin{adjustbox}{max width=\textwidth}
\begin{tabular}{lccc}
\toprule
Task & Grasp and Drop the Toy Tiger & Grasp and Drop the Blue Sponge & Put Blue Sponge to Right of Toy Tiger\\
\hline
Success Rate & 40\%  & 80\%  & 80\% \\
\bottomrule
\end{tabular}
 \end{adjustbox}
\caption{\reb{Experiments on three tasks using two unseen primitives. Success rates are calculated over 5 trials for each experiment.}}
\label{tab:prim}
\end{table}
}

\def\tabCotrain#1{
    \begin{table}[#1]
    \normalsize
    \centering
    \begin{tabular}{lccc}
    \toprule
                   & Pick and Place           & Poke                     & Average                  \\ \hline
    \algabbr (Co-train) & 13.3 ($\pm$5.8)          & 0.0 ($\pm$0.0)           & 6.7 ($\pm$3.0)          \\
    ICRT           & \textbf{65.0 ($\pm$7.3)} & \textbf{93.3 ($\pm$4.6)} & \textbf{79.2 ($\pm$4.6)} \\ \bottomrule
    \end{tabular}
    \caption{Ablation on co-training with DROID~\cite{khazatsky2024droid}. Training both DROID and ICRT-MT datasets in a single stage leads to worse task performance in both pick-and-place and poking tasks. Same as \cref{tab:results_table}, we evaluated each task primitive using six tasks not seen during training, conducting five trials per task for a total of 30 trials per primitive and 60 trials overall to calculate average performance. For each model, we report the mean success rate for each task, the overall success rate, and the corresponding standard error in parentheses.}
    \label{tab:ab_cotrain}
    \vspace{1mm}
    \end{table}
}

\def\tabfigMultiModal#1{
 \begin{minipage}{\textwidth}
  \begin{minipage}[b]{0.49\textwidth}
    \centering
    \adjustbox{max width=\textwidth}{
    \begin{tabular}{lccc}
    \hline
     & Orange       & Middle & Yellow       \\ \hline
        Goal Conditioned & 2/9          & 6/9    & 1/9          \\
        Prompt Orange    & \textbf{8/9} & 1/9    & 0/9          \\
        Prompt Yellow    & 0/9          & 1/9    & \textbf{8/9} \\ \hline
        \end{tabular}}
      \captionof{table}{We evaluate the multi-modal tiger-picking task by prompting the model with the goal, a trajectory that goes above the orange cup, and that goes above the yellow cup. We tally the number of times that the robot's passes above the yellow or the orange cup, or goes in between the two cups.}\label{tab:mmd_res}
    \end{minipage}
      \hfill
  \begin{minipage}[b]{0.49\textwidth}
    \centering
    \includegraphics[width=\textwidth]{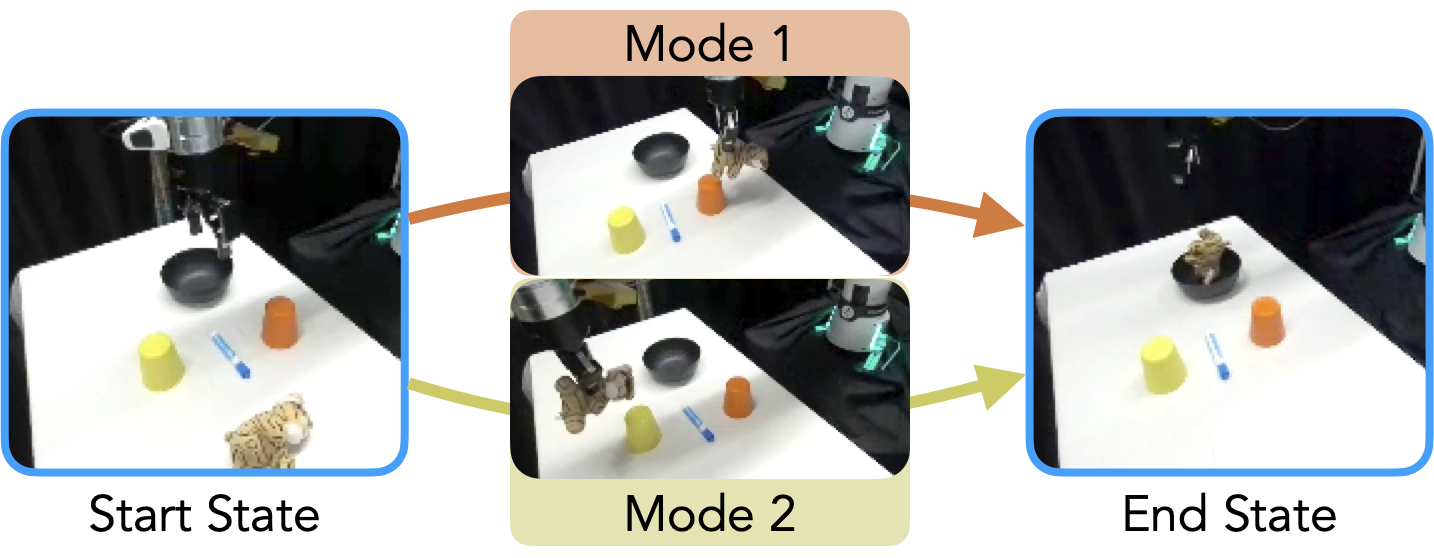}
    \captionof{figure}{An example of the multi-modal tiger-picking task, where there are two modes for the robot to reach to the goal state.}\label{fig:mmd-setup}
  \end{minipage}
  \end{minipage}
}

\def\tabPretrainHyperparams#1{
    \begin{table}[#1]
    \small
    \centering
    \begin{tabular}{cc}
    \toprule
    Config                 & Value                         \\ \hline
    optimizer              & AdamW                         \\
    base learning rate     & 1e-3                          \\
    learning rate schedule & cosine decay                  \\
    batch size             & 64                          \\
    weight decay           & 0.05                          \\
    optimizer momentum     & $\beta_1, \beta_2$ = 0.9, 0.999 \\
    warm up epoch          & 0.5                             \\
    total epochs           & 4                             \\
    proprioception noise   & 0.005                           \\
    action noise           & 0                               \\
    sequence length        & 512                             \\
    brightness augmentation& 0.1                             \\
    contrast augmentation  & 0.2                             \\
    num action prediction  & 16                             \\
    \bottomrule
    \end{tabular}
    \caption{Pre-training Hyperparameters}
    \label{tbl:pretrain_hyper}
    \end{table}
}

\def\tabFinetuneHyperparams#1{
    \begin{table}[#1]
    \small
    \centering
    \begin{tabular}{cc}
    \toprule
    Config                 & Value                         \\ \hline
    optimizer              & AdamW                         \\
    base learning rate     & 5e-4                          \\
    learning rate schedule & cosine decay                  \\
    batch size             & 64                          \\
    weight decay           & 0.01                          \\
    optimizer momentum     & $\beta_1, \beta_2$ = 0.9, 0.999 \\
    warm up epoch          & 1.25                             \\
    total epochs           & 125                             \\
    proprioception noise   & 0.005                           \\
    action noise           & 0                               \\
    sequence length        & 512                             \\
    brightness augmentation& 0.1                             \\
    contrast augmentation  & 0.2                             \\
    num action prediction  & 16                             \\
    \bottomrule
    \end{tabular}
    \caption{Finetuning Hyperparameters}
    \label{tbl:finetune_hyper}
    \end{table}
}

\def\tabInferenceFreq#1{
    \begin{table}[#1]
    \normalsize
    \centering
    \begin{tabular}{lc}
    \hline
                    & Inference Frequency \\ \hline
    \algabbr        & 39.6 Hz             \\
    \algabbr-Llama2 & 10.7 Hz             \\ \hline
    \end{tabular}
    \caption{Inference frequency of \algabbr, averaged over 100 steps. }
    \label{tbl:inf_freq}
    \end{table}
}

%% file: corl_2024/figures_tex/main_figures.tex
\def\figMethod#1{
    \begin{figure*}[#1]
        \centering
        \vspace{0.08in}
        \includegraphics[width=1.0\linewidth]{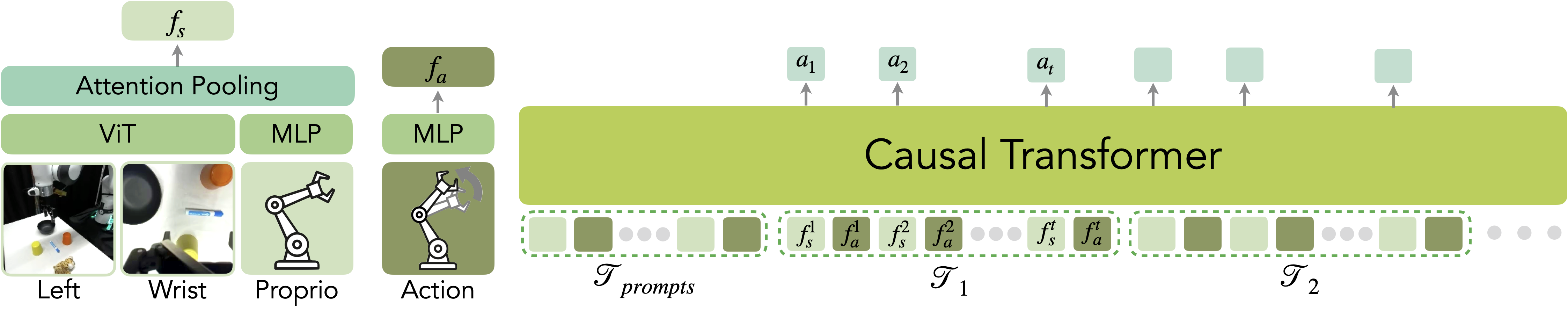}
        \caption{\textbf{Method Overview}: (Left) We encode camera observations with a pre-trained vision transformer. Additionally, we encode proprioception with an MLP. We concatenate the visual latent and the proprioception's latent and use attention pooling to extract a feature $f_s$ as the current state representation. We encode the current action with an MLP to get $f_a$. (Right) We concatenate multiple trajectories of the same task and randomly sample the first $k$ trajectories as the prompt. A causal transformer autoregressively predicts the next series of tokens. We decode the tokens that are at the position of the state features to generate the next $h=16$ action via an MLP. }
        \label{fig:method}
    \end{figure*}
}

\def\figInference#1{
    \begin{figure}[#1]
        \centering
        \includegraphics[width=0.98\linewidth]{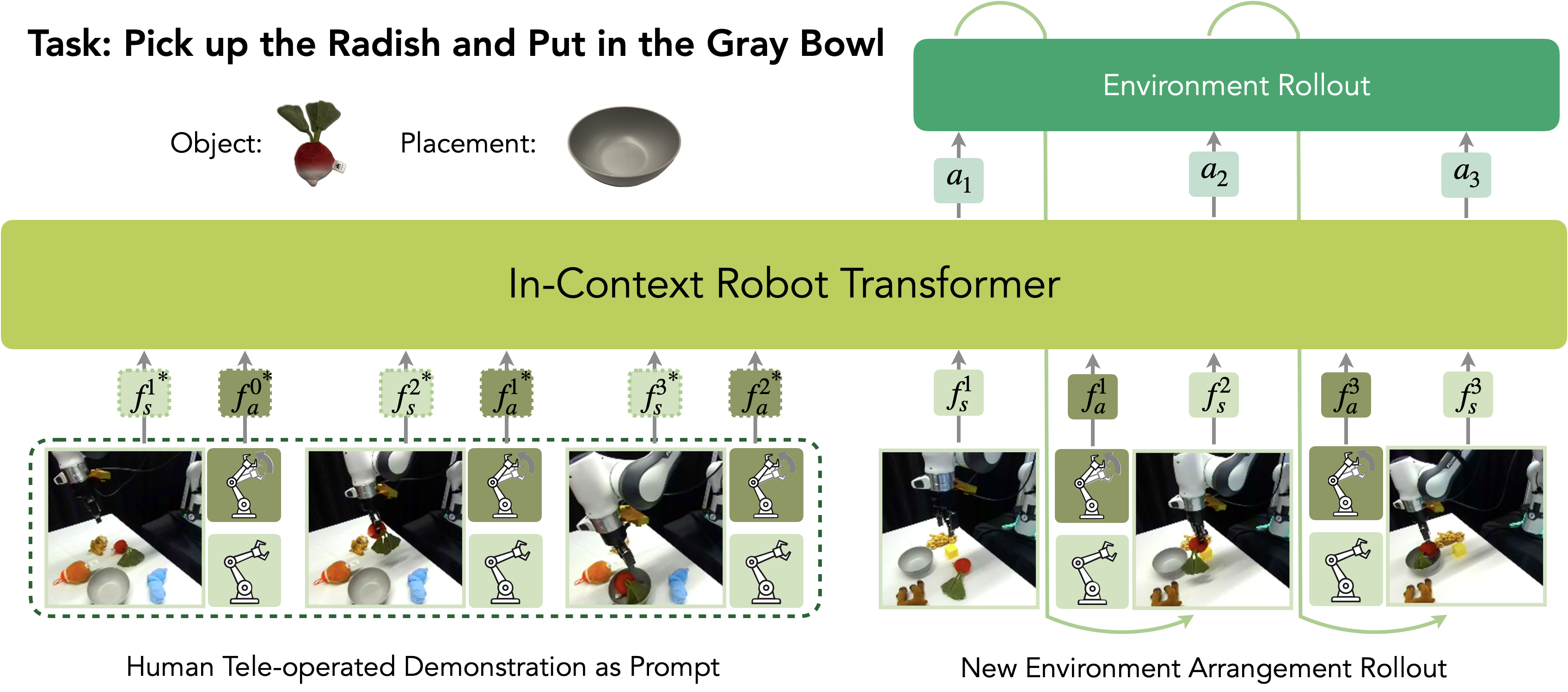}
        \caption{Example inference pipeline of \algabbr on the task of picking up the radish and putting in the gray bowl. A human teleoperated demonstration trajectory consisting of image observations, proprioception and actions are provided as the prompt. \algabbr takes the prompt and the current observation in a different environment to accomplish the task.}
        \label{fig:infer}
    \end{figure}
}

%% file: sections/0-abstract.tex
\begin{abstract}
We explore how to enable in-context learning capabilities of next-token prediction models for robotics, allowing the model to perform novel tasks by prompting it with human teleop demonstration examples without fine-tuning. We propose \algname (\algabbr), a causal transformer that performs autoregressive prediction on sensorimotor trajectories, which include images, proprioceptive states, and actions. This approach allows flexible and training-free execution of new tasks at test time, achieved by prompting the model with demonstration trajectories of the new task. Experiments with a Franka Emika robot demonstrate that the \algabbr can adapt to new tasks specified by prompts, even in environment configurations that differ from both the prompts and the training data. In a multi-task environment setup, \algabbr significantly outperforms current state-of-the-art robot foundation models on generalization to unseen tasks. Code, checkpoints and data are available on \url{https://icrt.dev}.

\end{abstract}

%% file: sections/1-intro.tex
\section{Introduction}

Learning-based single and multi-task robot policies have become increasingly capable~\cite{diffusion, zhao2023learning, lynch2023interactive, Reed2022, shah2023vint, roboagent, brohan2022rt, brohan2023rt2, chen2023palix, Driess2023}. This improvement in robot capabilities can largely be attributed to progress in related fields, particularly in vision and language modeling. \reb{Inspired by the recent development of large language models (LLMs) and large vision models (LVMs)~\cite{achiam2023gpt4, touvron2023llama, bai2023sequential}, which formulate natural language processing and vision problems all as next-token-prediction, recent works also have formulated robot learning as next-token-prediction problems and achieved state-of-the-art performance~\cite{brohan2022rt,brohan2023rt2,kim2024openvlaopensourcevisionlanguageactionmodel,octo_2023}.} Concurrently, there has been a surge in collecting large-scale robot datasets~\cite{depierre2018jacquard, Kalashnikov2018, Levine2018, acronym2020, shafiullah2023dobbe, fang2023rh20t, ebert2021bridge, walke2023bridgedata} and pre-training models on these datasets~\cite{nair2022, xiao2022, ma2022vip, radosavovic2022, octo_2023}. 

\reb{Despite being pre-trained on large datasets and showing some generalization ability, it is still challenging to teach these models to perform unseen tasks in different environments without additional training.} New human demonstrations via teleoperation or new data collected from hand-crafted motion primitives, as well as another round of model-finetuning, are often needed to complete the new tasks. This process adds complexity to the workflow, making it challenging to apply these methods in real-world environments. Ideally, given one or a few demonstrations, the robot should be able to perform the task \textit{immediately}. In their respective domains, LLMs and LVMs~\cite{achiam2023gpt4, touvron2023llama, bai2023sequential} have exhibited a similar ability, named \textit{in-context learning}: a capability allowing the model to rapidly adapt to and recognize the task corresponding to the prompt provided at inference time without additional training. 

\reb{\textit{Is the in-context learning capability of next-token prediction models limited to vision and language domains?}} In this paper, we introduce~\algname (\algabbr), where we explore how next-token prediction models can be extended to perform real-robot in-context learning. \reb{For~\algabbr, the context is provided as a series of robot trajectories corresponding to a new task. The model learns from this context to perform the task in a different environment configuration without requiring additional training. A robot trajectory is a sequence of image observations, robot proprioceptive states, and actions. This trajectory implicitly encodes task primitives and the objects the robot needs to interact with. The model extracts this information from the prompt and then executes actions following a similar pattern in its current environment.

}

\reb{
Compared to existing few-shot imitation learning approaches, \algabbr offers a simple framework that avoids complicated loss functions, prior knowledge, and the need to identify key points or key frames, and operates directly on raw robot trajectories for continuous control. Additionally, unlike existing next-token prediction models for robot learning, \algabbr features a long context window, allowing it to train on multiple sensorimotor trajectories from the same task and use one or more sensorimotor trajectories as prompts during inference.

Importantly, we observe that certain properties of the dataset are crucial for enabling in-context learning on real robots. Specifically, datasets that allow multiple tasks to be performed from the same initial observation are particularly beneficial. Unlike existing single-task datasets or many multi-task datasets where each environment has a unique object for robot interaction, these scenarios require the model to rely on the prompt to correctly identify the task and determine the appropriate object for interaction.
}

We make the following contributions:
\begin{enumerate}
    \item We introduce \algabbr, a robot foundation model that performs in-context learning on a real robot, which can effectively learn from context trajectories and perform unseen tasks without training.
    \item We provide a new multi-task robot dataset and a training paradigm for fostering multi-task and in-context capability at inference time.  
    \item Physical experiments on a Franka Emika robot demonstrate that \algabbr can learn from the provided context and perform the unseen tasks specified by the prompt at various generalization levels.
\end{enumerate}

%% file: sections/2-related-works.tex
\section{Related Works}

\subsection{Multi-Task Imitation Learning for Robotics}

\reb{
Imitation learning is an effective paradigm for equipping robots with various skills. The simplest algorithm in this domain, behavior cloning, has been successful across a wide range of tasks~\cite{pomerleau, argall2009survey, Levine2016}. In recent years, alternative architectures such as energy-based models~\cite{Florence2021ImplicitBC} and diffusion models~\cite{diffusion} have also been proposed. Typically, these approaches require training a \textit{separate} model for each task, although multi-task policies can be distilled from these task-specific models after training~\cite{ha2023scaling}.

Recent advancements have shown that using transformers for next-token prediction in sequence modeling has been particularly effective in both language and vision domains, especially for \textit{multi-task learning}~\cite{Brown2020,touvron2023llama,liu2023llava}. In pursuit of developing generalist agents and multi-task robot policies, robot action planning is framed as a next-token prediction task using transformer-based architectures trained on large multi-task robot datasets~\cite{Rpt2023, radosavovic2024humanoid, brohan2022rt,brohan2023rt2, kim2024openvlaopensourcevisionlanguageactionmodel, octo_2023, jang2022bc, jiang2022vima, reed2022generalist, 2024rtx, shah2023vint}. 
Octo~\cite{octo_2023} and OpenVLA~\cite{kim2024openvlaopensourcevisionlanguageactionmodel} represent the state-of-the-art among multi-task robotic policies. Octo~\cite{octo_2023} conditions on both goal images and language instructions, utilizing a transformer architecture with a diffusion head that fuses these inputs with current image observations to predict robot actions. OpenVLA\cite{kim2024openvlaopensourcevisionlanguageactionmodel} conditions solely on language instructions, fine-tuning a pre-trained vision-language model to predict robot actions based on visual observations and language inputs.

}

\subsection{In-Context Learning}

Despite training on large datasets, multi-task policies often struggle when faced with new objects, tasks, or environments, frequently requiring fine-tuning. Meta-learning has been shown to increase fine-tuning efficiency for generalization to new tasks~\cite{finn2017model, finn2017one, xu2023hyper}, which has led to progress in few-shot imitation learning. To simplify the application of learned policies in the real world, recent approaches focus on methods that avoid fine-tuning model parameters for task generalization. Instead, these methods teach the model by providing demonstrations of tasks~\cite{di2024keypoint, jain2024vid2robot}.~\citet{Brown2020} refers to this as ``in-context learning", distinguishing it from approaches that rely on parameter fine-tuning.

Many in-context learning methods often employ contrastive learning to train context encoders, which identify the most similar training tasks to the test task in the latent space~\cite{jang2022bc,mandi2022generalizableoneshotvisualimitation}. However, how to effectively integrate these methods within the next-token-prediction framework remains unclear. \citet{valassakis2022demonstrate} achieved one-shot in-context learning by training a visual servoing network to align the robot's end-effector with the object's relative pose during the demonstration, but this approach requires an additional object segmentation model. \citet{di2024keypoint} introduced Keypoint Action Tokens, demonstrating in-context imitation learning using a large language model by representing demonstration trajectories as 3D coordinates with few-shot prompting. Unlike these approaches, \algabbr operates without additional perception modules, processing raw image observations directly. Additionally, Vid2Robot~\cite{jain2024vid2robot} developed an encoder-decoder transformer that uses a demonstration video of a human and the current robot state as the prompt to generate robot actions. However, this method requires many auxiliary losses while \algabbr uses a simple next-token prediction loss.

In this paper, we focus on enhancing next-token-prediction models to perform real-world in-context imitation learning with robots. \algabbr bypasses the need for additional context encoders by directly using robot sensorimotor trajectories from new tasks as prompts for the transformer-based model. \algabbr is closely related to the seminal work, One-Shot Imitation Learning~\cite{duan2017one} and Prompting Decision Transformer~\cite{xu2022prompting}.~\cite{duan2017one} predicts the next action by applying cross-attention between a demonstration sequence on a new task and the current environment's state, while~\cite{xu2022prompting} employs a short trajectory prompt to encode task-specific information for guiding policy generation in offline reinforcement learning, using full state information and known reward functions. While both of these methods show their effectiveness in simulation, it is hard to have full-state information and known reward functions for all real-world robot manipulation tasks. To address these challenges, \algabbr does not model rewards, utilizes a significantly longer context window, and demonstrates in-context learning capabilities in physical experiments using image observations.

%% file: sections/3-problem-statement.tex
\begin{figure*}
    \centering
    \includegraphics[width=\textwidth]{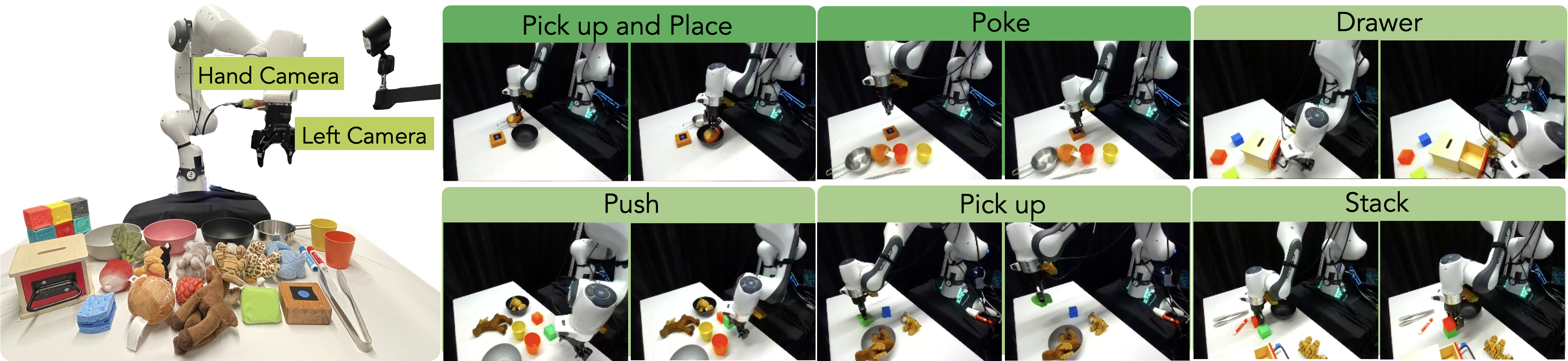}
    \caption{Our physical setup with the Franka Emika robot, the wrist and side camera and the objects used in training and evaluation. We consider 6 primitives for training and choose ``pick up and place'' and ``poke'' as the primitives for evaluation (dark green).}
    \label{fig:phy-setup}
\end{figure*}
\section{Problem Statement}
\label{sec:problem_statement}
We consider in-context imitation learning under a real-robot manipulation setting. The goal is to train a model with in-context learning capabilities using a multi-task robotic dataset. At test time, the model can handle unseen tasks in novel environment configurations by using a few new human-teleoperated robot demonstrations as prompts. Here, environment configuration refers to the objects present in the scene and their spatial arrangement. Notably, this process is achieved \textit{without any additional training} on the new demonstrations.

We define \textit{motion primitives} as distinct robot actions utilized to accomplish various \textit{tasks}. Each task is characterized by 1) a specific motion primitive and 2) the set of objects the robot interacts with using that primitive. By altering the environment configuration at test time compared to the one in the prompt, we assess the model's ability to select the appropriate motion primitive and identify the correct object for interaction. In this work, we consider new tasks to be tasks involving unseen objects but using motion primitives from the training data (for example, training on picking up a tiger toy and testing on picking up a cube).

We make the following assumptions for \algabbr experiments:
\begin{enumerate}
    \item The model is trained on a diverse multi-task demonstration dataset. Each trajectory contains RGB observations from a fixed camera and a wrist-mounted camera, proprioception and action.
    \item The task tested on the robot is within the reachable workspace of the robot.
\end{enumerate}

%% file: sections/4-method.tex
\figMethod{t!}

\section{Approach}
In this section, we first introduce the data composition to facilitate in-context imitation learning. We then introduce the architecture and training objective for the transformer-based policy to effectively leverage the data. 

\subsection{Data Formulation}\label{ssec:data}

For model training, we consider a dataset $\mathcal{D}$ of visuomotor trajectories $\mathcal{T}$. Each trajectory of length $t$ is a sequence of camera images $i_t$, proprioceptive robot states $s_t$, and actions $a_t$: $\mathcal{T} = (i_1, s_1, a_1, ..., i_t, s_t, a_t)$. \reb{We use the absolute end-effector pose as the robot's proprioceptive state and the delta robot end-effector pose between time steps as the action, which consists of delta translation, delta rotation and the continuous gripper action.} We assume a known grouping of the trajectories so that the dataset can be partitioned into disjoint sets of tasks $\mathcal{D} = \bigcup_{k=1}^K S_k$, with $S_k \cap S_\ell = \emptyset, \ k \neq \ell$, where $\mathcal{S}_k = \{\mathcal{T}_{k_1}, ..., \mathcal{T}_{k_n}\}$. In practice, this grouping can be retrieved from the semantic labels of the dataset. In this work, we utilize the existing large robotic dataset DROID~\cite{khazatsky2024droid} and a multi-task dataset manually collected in our robot setup, which we name \algabbr-Multi-Task (\algabbr-MT).

DROID~\cite{khazatsky2024droid} is a joint effort from different organizations and contains 76k real-world demonstrations. We randomly sample 10k demonstrations from DROID after filtering out demonstrations shorter than 30 steps and longer than 450 steps. DROID dataset labels the task through human-specified language instructions, which may be different for the same task. We organized the DROID data by grouping demonstrations based on their language instructions CLIP text embedding cosine similarity. Specifically, we use a threshold of 0.9 for grouping demonstrations. To further facilitate in-context learning, we make sure that each task group contains at least 4 trajectories so that there are sufficient trajectories to serve as prompts for each other. This results in roughly 2k trajectories that we use for pre-training \algabbr.

Many trajectories in the DROID dataset are collected in a single-task setup, where only one task can be completed in a scene (e.g. only one object is presented). In such a setup, the model can learn a shortcut solution to perform the task, by only focusing on the current observation but not the prompt. Therefore, multi-task data is crucial for the model to learn from the prompt. We manually collected a multi-task dataset \algabbr-Multi-Task (\algabbr-MT) using the DROID setup (Figure.~\ref{fig:phy-setup}). This dataset has 1098 trajectories in total, and contains 29 tasks with 6 primitives: picking, pick-and-place, stacking, pushing, poking, opening and closing drawers. Objects used in the data collection and examples of the primitives are shown in Figure.~\ref{fig:phy-setup}. In \algabbr-MT, each environment is set so that there exist more than 2 possible tasks for the current observation so that the model has to distinguish and learn the motion from the prompt. 

During the training, for each trajectory, we independently apply vision augmentation on the image observations by augmenting the brightness and contrast. We additionally apply random crops and scaling to the side camera observation. We also apply proprioception noise sampled from a normal Gaussian distribution $\mathcal{N}(0, 0.005)$. For each epoch, we randomly shuffle the order of trajectories from each task and concatenate them to form the training sequence. For each batch, we sample a subsequence of length $L=512$ as the input to the model, where $L$ is the sequence length defined as the number of observation, state, and action tuples. In practice, 512 steps usually contain up to 5 trajectories from the same task. We randomly select the first $k$ trajectories and label them as the prompt within the sequence. At least one complete trajectory is included in the prompt. 
This data grouping aims to capture inter-trajectory patterns, encouraging the model to generate action conditioned on the prompt trajectories. This approach differs from traditional behavior cloning methods, which typically use short input sequences that focus on modeling intra-trajectory behaviors.

\subsection{Model Architecture}

We construct the \algabbr model with three parts: a pre-trained vision encoder, a series of projectors for each input modality, and a causal transformer backbone (Figure~\ref{fig:method}). 

\textbf{Vision Encoder} The model processes multi-view image observations through a pre-trained vision transformer. However, most visual pre-trained networks are trained on ImageNet or human videos~\cite{radosavovic2022, vc2023, chen2024sugar, nair2022}, which exhibit a significant domain gap when compared to typical images from robot datasets, where the images frequently include robots or grippers. To minimize the domain gap, we pre-train a vision transformer~\cite{Dosovitskiy2020} (ViT-Base) on an equal mix of ImageNet~\cite{Deng2009} and Open X-Embodiment~\cite{2024rtx} data, using CrossMAE as an efficient pre-training method~\cite{fu2024rethinking}.
During the training of the \algabbr model, we freeze the vision encoder for efficiency. The vision encoder outputs the entire feature map for each of the cameras and is then fed into the proprioception projector (\cref{fig:method} left).

\reb{\textbf{Modality-Specific Projectors} To project image observations, the robot's proprioceptive state, and actions into a shared latent space for sequence modeling, we design modality-specific projectors.} At each timestep, the model takes as input a token representing either an observation or an action. To produce a single state token that captures fine-grained visual information and the proprioceptive state of the robot, we use attention pooling~\cite{lee2019set} between all visual tokens from a single camera's observation and a proprioception embedding produced by a multi-layer perceptron (MLP). The resulting embeddings for each camera are concatenated to produce a single state token $f_s^t$ of dimension equal to the transformer latent dimension. Similar to proprioception, the action is embedded with an MLP into an action token $f_a^t$. This process produces a sequence of state and action tokens that are passed into the transformer.

\textbf{Transformer Model} The encoded sequence of state and actions is passed into a Transformer model~\cite{Vaswani2017}, following the design of Llama2~\cite{touvron2023llama}. The transformer takes as input the sequence of state and action features $(f_s^1, f_a^1, \cdots, f_s^t, f_a^t)$ that are produced by the modality-specific projectors. We add MLP decoders to produce state and action outputs from the last layer of the transformer at the appropriate positions. We denote the transformer with the decoder heads as $g_\theta$. Therefore, the desired outputs are the shifted sequence of proprioceptive states and actions $(a^1, s^2, a^2, \cdots, a^t, s^{t+1})$. This naturally forms a next token prediction problem, as $g_\theta(f_s^1)$ predicts $a^1$ and $g_\theta(f_s^1, f_a^1, \cdots, f_s^n)$ predicts $a^{n+1}$. In practice, we find it beneficial to predict the next $h$ actions at each time step, and use temporal ensembling~\cite{zhao2023learning} to execute the final action.

\reb{Inspired by Octo~\cite{octo_2023} and vision transformers~\cite{Dosovitskiy2020}, we consider a randomly initialized Llama2 model of 12 layers with a latent dimension of 768, which we name \textit{Llama2-Base}.} In addition, multiple works have shown that multimodal inputs can be aligned to large-language models~\cite{liu2023llava, han2023imagebindllm, fu2024tvl, brohan2023rt2, mirchandani2023large}. Multi-modal language model, Palm-E~\cite{Driess2023} has shown success in enhancing generalization when being directly incorporated into robotic control~\cite{brohan2023rt2}. Therefore, we also investigate the effectiveness of using a large-language model for in-context robot learning by initializing the transformer with a pre-trained Llama2-7B. Due to the large domain gap between natural language and robot trajectories, a frozen language model may not be sufficient. Therefore, similar to prior work in multimodal alignment, we fine-tune the language model with LoRA~\cite{hu2021lora}, with a rank of 32. Due to compute resource limitations, we are unable to fully fine-tune the model.

\textbf{Loss Function} 
To provide more supervision signals so that the model can better respond to the trajectory ``prompt'' we provide at test time, we reference works in training multi-turn conversation chatbots~\cite{chiang2023vicuna, liu2023llava}, where they only compute loss on the response generated by the chatbot, instead of the prompt. Recall that in \cref{ssec:data}, we randomly sample the subsequence of the concatenated trajectories as the prompt trajectory. Analogously, we only compute action prediction with L1-loss for the actions after the prompt trajectories.

\textbf{Inference} The simplicity of the next-token prediction objective makes inferencing with \algabbr straightforward at test time. As shown in Figure.~\ref{fig:infer}, we provide one or more human-teleoperated demonstrations in the form of robot sensorimotor trajectories (formatted identically to the training data), along with the current image observations and the robot's proprioceptive state as inputs. The model then predicts the next action, which is executed by the robot. After each action, the policy receives updated image observations and proprioceptive state, allowing it to iteratively predict and execute subsequent actions.

A key advantage of this framework is its use of the transformer's sequential processing capability. Instead of reprocessing the entire sequence history for each model evaluation, as seen in previous works~\cite{octo_2023, kim2024openvlaopensourcevisionlanguageactionmodel, brohan2022rt, brohan2023rt2}, the model employs a key-value (KV) caching mechanism, as discussed in~\cite{touvron2023llama}. This mechanism stores previous outputs, allowing the model to compute only the outputs for the new token. This approach significantly reduces computational overhead, lowering the complexity from quadratic to linear relative to the sequence length. Empirically, \algabbr can inference at 39.6 Hz, allowing it to perform real-time close-loop control.

%% file: sections/5-experiments.tex
\section{Experiments}
\label{sec:experiments}
\figInference{t}
\reb{
In this section, we design an experimental setup to evaluate the in-context learning capabilities of the proposed models and compare them against several baselines. Instead of focusing on the difficulty of learning a specific task primitive, we design the experiments to assess the policy's ability to accomplish novel tasks based on the provided prompt trajectories.%

\textbf{Experiment Design} We consider two action primitives: a \textit{pick-and-place} primitive and a \textit{poking} primitive. For each action primitive, we design \textit{six unseen tasks} (as defined in Section~\ref{sec:problem_statement}), with three tasks evaluating \textit{in-domain} object generalization (selected from \textit{yellow cube, red cube, black cube, pink bowl, and blue bear} and three objects \textit{unseen} during training (selected from \textit{radish, blue sponge, grey dog, and black dog}). 

Each task has five difficulty tiers. In the pick-and-place task, the model must identify the correct object to grasp and where to place it in a multi-object or multi-placement scenario. The tiers include: 1) no distractors, 2) one distractor object, 3) two distractors, 4) three distractors, and 5) one distractor placement position. In the poking task, the robot closes the gripper, pokes the object, lifts the end-effector, and opens the gripper, with tiers involving 0-4 distractor objects.

The pick-and-place task is scored with 0.5 for a correct pick and 1 for a successful placement. In the poking task, failure is marked if the wrong object is poked. The model has 25 seconds (375 steps) for retries. Each difficulty level is attempted once, and we report the average success rate per task, along with the average success rate and standard deviation across the six tasks for each action primitive.

\textbf{Models} The default \textbf{\algabbr} is a randomly initialized Llama2-Base model pretrained on DROID and fully fine-tuned on \algabbr-MT. We evaluate the impact of model initialization and training datasets by introducing the following three variants: 1) \textbf{\algabbr-Llama2}, a pre-trained Llama2-7B language model fine-tuned on \algabbr-MT with LoRA; 2) \textbf{\algabbr (DROID)}, a randomly initialized Llama2-Base model trained only on the DROID dataset; and 3) \textbf{\algabbr (MT)}, a randomly initialized Llama2-Base model trained only on the \algabbr-MT dataset.

We consider 3 baseline models. We train a goal-conditioned policy, where the goal observations are always prepended to the sequence, and each sequence is from one trajectory. This resembles the normal goal-conditioned imitation learning setup. Additionally, we finetune Octo~\cite{octo_2023}, the state-of-the-art goal-image and language conditioned policy, and OpenVLA~\cite{kim2024openvlaopensourcevisionlanguageactionmodel}, the state-of-the-art language conditioned multi-task imitation learning policy. Octo is fine-tuned using their official fine-tuning recipe. We incorporate action chunking into OpenVLA by asking it to predict the next 16 actions, which performs better than vanilla OpenVLA which predicts only the next step. Both of these methods are representative of robot policies that use next-token prediction objectives.

\textbf{Prompt Generation} For each task, we collect 3 demonstrations (with zero, one distractor object, a distractor placement for pick-and-place, or two distractor objects for poking) as the prompt in total before running the experiment. Please refer to the \href{http://icrt.dev}{website} for a visual example. During testing, a random demonstration is drawn as a prompt to assess the model's ability to generalize to different prompts. It's important to note that the environment setup during policy rollout \textit{differs} from the prompts' setup, ensuring that the evaluation measures the model's understanding of task-relevant information from the prompt, rather than simply copying actions from it.

\tabResultsTable{tbp!}
\textbf{Results} We present the results in \cref{tab:results_table}. For the pick-and-place primitive, we observe that the goal-conditioned policy generally succeeds in identifying the correct object to grasp when no distractor objects are present. However, its performance degrades significantly as the number of distractors increases. When the goal image only specifies the task but not the specific way to achieve it in the current environment, goal-conditioned policies often fail to execute the task effectively.

Octo struggles with determining which object to interact with and where it should be placed, highlighting the challenges posed by our experimental setup for multi-task policies. OpenVLA, while often moving towards the correct object, frequently fails in grasping the object or mistakenly performs the wrong task (e.g., grasping instead of poking, and vice versa). This indicates that OpenVLA may require a greater number of demonstrations (more than 50) per task to achieve better performance, and that relying solely on language conditioning may not be sufficient for generalization to new tasks.

The results suggest that \algabbr outperforms the goal-conditioned policy in identifying the correct object to pick up and the appropriate placement location. The poking task presents a significant challenge for the goal-conditioned policies, as the goal position often closely resembles the start configuration. However, after conditioning on the prompt trajectory, \algabbr is able to correctly identify the task as poking, and the results indicate that it consistently reaches the correct target object while ignoring distractors. Despite this, we do observe some failure modes with \algabbr, such as missing the grasp of the target object, grasping the wrong object, or placing objects in incorrect locations. Specifically, when a distractor object shares the same color but has a different shape, the model struggles to accurately determine which object to grasp. This implies that additional fine-tuning of the vision encoder might be required to enhance model performance, a conclusion also reached by OpenVLA~\cite{kim2024openvlaopensourcevisionlanguageactionmodel}.

\tabAblationTable{tbp!}

\section{Ablations}
In this section, we provide additional experiments presented \cref{tab:ablation_table} that ablate on a few core design choices. We provide additional ablation studies on the \href{https://icrt.dev}{official website}.

\subsection{Model Initialization}
We conducted ablation studies to examine the impact of using a pretrained Llama2 on language data and fine-tune it for robot sensorimotor sequence modeling. The results, presented in \cref{tab:ablation_table}, show that although \algabbr-Llama2-7B achieves a lower training loss, its performance is worse compared to its smaller counterparts. This discrepancy may be attributed to a lower inference frequency of \algabbr-Llama2 (10.7 Hz vs 39.6 Hz). Future work can focus on optimizing the inference speed of \algabbr-Llama2 to improve performance.

\subsection{Training Dataset}
\reb{We find that training on the DROID subset (see \cref{ssec:data}) is insufficient for completing any of the test tasks; the policy (\algabbr (DROID)) shows no progress across all tasks. This suggests that although the DROID subset may offer greater visual diversity, the unique structure of \algabbr-MT—where multiple tasks are performed from the same initial observation—is particularly beneficial in developing the in-context learning capabilities of a next-token prediction robot model.}

\algabbr (MT) shows similar performance to \algabbr that is pre-trained on DROID, especially for the pick-up and place primitive, even surpassing \algabbr on the \textit{put radish in grey bowl} task. However, \algabbr (MT) does not perform as well on the poking primitive. The results suggest that it may be beneficial to pre-train the autoregressive model on a large dataset, as a diverse dataset may help the transformer to perform better alignment between visual features and control.

\subsection{No Prompt Loss}
Following the design of many multi-turn conversation large language models or vision language model fine-tuning works~\cite{liu2023llava, chiang2023vicuna, liu2023improvedllava, dai2024instructblip}, we do not calculate the loss for the predicted action in the prompt trajectories but only do so on the predictions after the prompt trajectories. We mark the model that calculates loss on the prompt as \textbf{\algabbr+Prompt Loss} and the default model as \textbf{\algabbr}. The results are shown in \cref{tab:ablation_table}. We find that only predicting the trajectories after the designated prompt trajectories can significantly improve the model's performance. We hypothesize that in the situation where there is a loss on the prompt trajectories, the model is forced to do unconditional generation based on current observations for those prompts. This may cause the model to stop paying attention to the prompt, especially when there are multiple possible tasks available.

}

%% file: sections/6-conclusion.tex
\section{Limitations and Conclusion}
While results suggest that~\algabbr learns from the prompt trajectories and generalizes to unseen objects, tasks, and certain primitives that resemble the ones in training\footnote{Please refer to the appendix and videos on \href{https://icrt.dev}{the website} for more detail.}, it is still unclear how to generalize to completely unseen action primitives. Future works should investigate how scaling model capacity and scaling dataset can help with primitive-level generalization. In addition, \algabbr assumes a fixed robot morphology with a fixed impedance controller. Future works can also investigate how to facilitate transfer between different robot morphologies by learning a unified policy on different robots. \algabbr-Llama2 has a low inference frequency which may contribute to its low performance. We hope to speed up \algabbr-Llama2 at inference time in the future.

In summary, we present \algabbr, where we study in-context, multi-task imitation learning on a real robot. We achieve this by training a causal transformer model on sequences of robot trajectories, where trajectories from the same task are combined to provide context for task execution. Additionally, we introduce a multi-task dataset to facilitate this in-context learning approach. Our experiments show that by using robot sensorimotor trajectories as context, the model can generalize learned motion primitives to unseen objects and novel environment configurations, particularly in scenarios where multiple tasks are present.

%% file: sections/8-acknowledgement.tex
\section{Acknowledgement}
This research was performed at the AUTOLAB at UC Berkeley in affiliation with the Berkeley AI Research (BAIR) Lab, and the CITRIS "People and Robots" (CPAR) Initiative.
In their academic roles at UC Berkeley, Letian Fu, Huang Huang, Gaurav Datta, Lawrence Yunliang Chen, William Chung-Ho Panitch, Fangchen Liu, and Ken Goldberg are supported in part by donations from Autodesk, Meta, Google, Siemens, Toyota Research Institute, Bosch, and by equipment grants from PhotoNeo, Nvidia, NSF AI4OPT Centre, and Intuitive Surgical. L.Y. Chen is also supported by the National Science Foundation (NSF) Graduate Research Fellowship Program under Grant No. 2146752. We thank Jiayi Pan, Xinyang Geng, Dantong Niu, and Chung Min Kim for their helpful discussions and feedback.

%% file: sections/7-supplement.tex
\newpage 
\onecolumn
\section{Supplementary Material}
\subsection{Scene Illustrations}\label{apx:scene}
\begin{figure}[h!]
    \centering
    \includegraphics[width=0.9\linewidth]{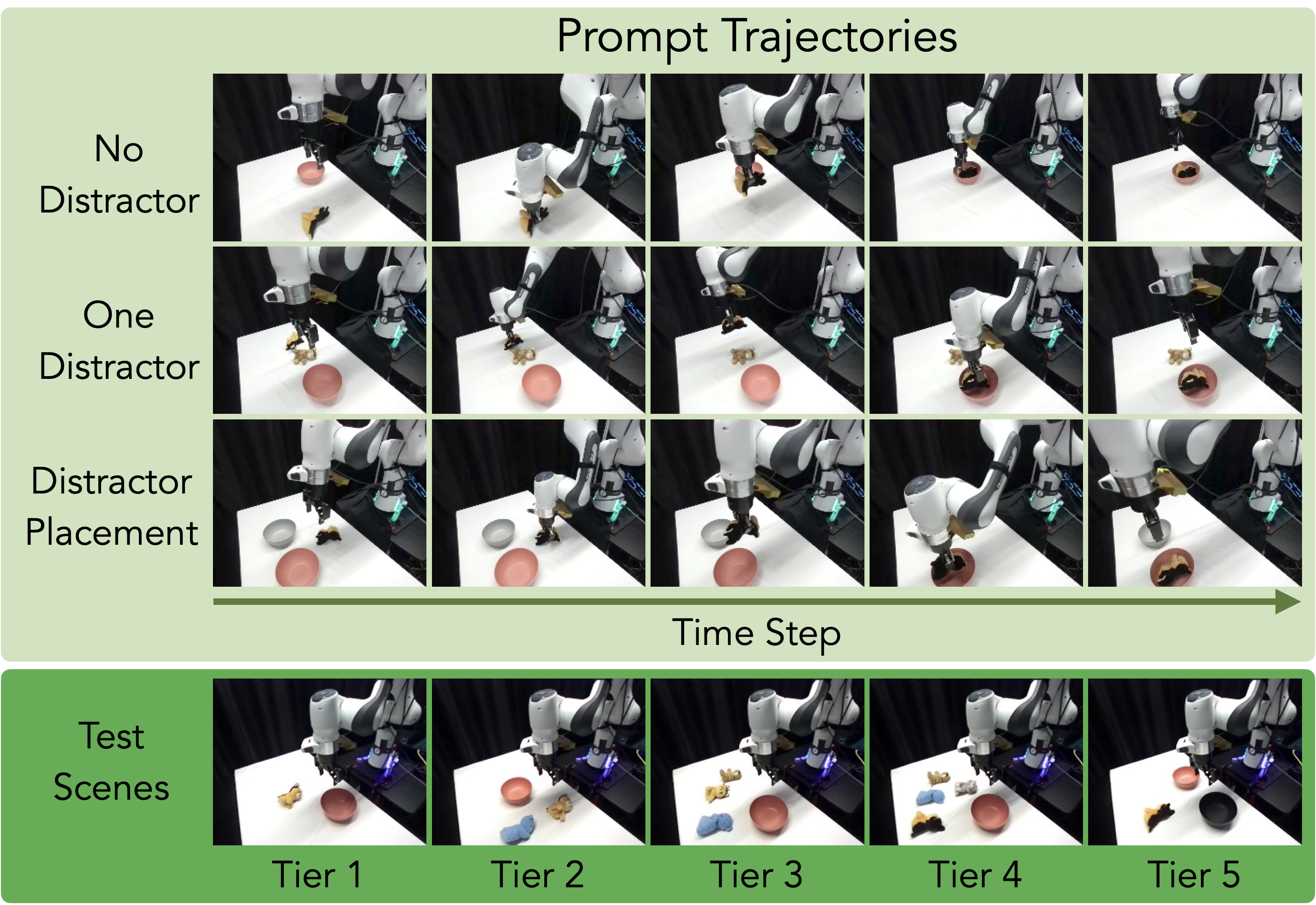}
    \caption{\reb{Illustrations of the prompt trajectories (top) and test scenes (bottom) for the pick up the black dog and place in the pink bowl task. Three prompt trajectories of different types are collected. The test scenes are different from all prompt trajectories and 5 tiers of scenes with different number of distractors are considered. }}
    \vspace{-10pt}
    \label{fig:prompt-testscene}
\end{figure}

\reb{We provide illustrations on the prompt trajectories and test scenes for the pick up the black dog and place in the pink bowl task in Figure~\ref{fig:prompt-testscene}. As mentioned in Section~\ref{sec:experiments}, we collected 3 types of prompt trajectories and test \algabbr on 5 tiers of scenes that are different from the scenes in the prompt trajectories.}

\subsection{Ablation Studies}\label{ssec:appendix_ablations}
In this section, we provide additional ablation experiments on a few core design choices and different prompting strategies.

\reb{
\subsubsection{Repeatability Experiments}
We conduct experiments to evaluate the repeatability of the performance of \algabbr. We conduct a pick up the black dog and place in the pink bowl task and a poke blue sponge task for 5 rollouts, where each rollout contains 5 trials as in Section~\ref{sec:experiments}, resulting a total of 25 trials. We calculate the average and the standard deviation of the success rate. Results are shown in Table~\ref{tab:std}. The low std from Table~\ref{tab:std} suggests that the \algabbr can reliably achieve the task.
\tabStd{h!}
}

\reb{
\tabPrompt{h!}
\subsubsection{Prompt Trajectories}
We conduct experiments on different prompt types to evaluate the effect of different prompt trajectories on task performance. We consider the task of picking up a black dog and placing in a pink bowl. We have three prompt trajectories of different types: one with no distractors, one with one distractor and one with one distractor placement, as shown in Appendix Figure~\ref{fig:prompt-testscene} top. All three prompts trajectories are collected by human teleoprating the robot. The object locations and the placement locations at test time are different from that in all three prompts. As in Section~\ref{sec:experiments}, for each prompt type, we conduct the task with 5 trials as shown in Appendix Figure~\ref{fig:prompt-testscene} bottom. The average success rates are reported in Table~\ref{tab:prompt}. We conduct experiments with one prompt trajectory of different types (the first three columns in Table~\ref{tab:prompt}), two prompt trajectories and three prompt trajectories. All prompt types result in similar performance, indicating \algabbr is not sensitive to the prompt trajectory types. We hypothesize this is because during the training, \algabbr has seen different types and numbers of prompts.

}

\reb{
\subsubsection{Unseen Primitives}\label{ssec:unseen_primitives}
\tabPrimitive{h!}
We evaluate the generalization capability of \algabbr on primitives that are unseen during the training but resemble the training primitives. We consider two such unseen primitives: grasp and drop an object and put object A to the right of object B. We consider three tasks: grasp and drop a toy tiger, grasp and drop a blue sponge (unseen objects during training) and put the blue sponge to the right of the toy tiger. As in Section~\ref{sec:experiments}, we conduct 5 tiers for each task. Experiment results are summarized in Table~\ref{tab:prim}, where \algabbr shows decent success rate on all three tasks, suggesting that \algabbr can generalize to some unseen primitives that resemble the training primitives.
}

\subsubsection{Co-training}
For training \algabbr, we opt to separate the training into two stages: a pre-training phase where the model is pre-trained on the DROID dataset~\cite{khazatsky2024droid}, and a fine-tuning phase where the model is trained on the ICIL-MT dataset. In this ablation, we experiment with whether these two can be combined into a single stage, where the policy is end-to-end trained with DROID and ICIL-MT. To balance the two datasets, we first calculate the median number of trajectories per task across the two datasets, then for each epoch, sample each task with the median number of trajectories. This allows each task to be equally represented in each epoch. We train the model for the same number of epochs as for \algabbr fine-tuning and report the results in \cref{tab:ab_cotrain}. The results indicate that the model does not converge as quickly in the combined stage and fails to respond to prompts and complete tasks effectively. We hypothesize two reasons for this: firstly, the dataset is heavily biased towards DROID, which contains 200 tasks compared to only 29 tasks in ICIL-MT, making it difficult for the model to learn the tasks as effectively as in the separate stage training. Future works can analyze the data mixture and how to train with large-scale datasets more effectively.

\tabCotrain{h!}

\subsection{Detailed results}
In this section, we present the per-task performance for the pick-and-place primitive (\cref{tab:pick_place}) and the poking primitive (\cref{tab:poke}). For each action primitive, we design \textit{six unseen tasks} (as defined in Section~\ref{sec:problem_statement}), with three tasks evaluating \textit{in-domain} object generalization (selected from \textit{yellow cube, red cube, black cube, pink bowl, and blue bear} and three objects \textit{unseen} during training (selected from \textit{radish, blue sponge, grey dog, and black dog}). 
\tabPickPlace{h!}
\tabPoke{h!}

\subsection{Hyperparameters}
We provide the hyperparameters for both the pre-training and fine-tuning phase in~\cref{tbl:pretrain_hyper} and~\cref{tbl:finetune_hyper}. 
\tabPretrainHyperparams{h!}
\tabFinetuneHyperparams{h!}

\subsection{Parameterization}
\label{ssec:parametrization}
\textbf{Proprioception} The proprioception space is parameterized by the absolute end effector translation (x, y, z), a 6DoF rotation vector, and a continuous end-effector gripper state. This results in a 10-dimensional proprioception representation. The 6DoF rotation vector is flattened from the $\textit{SO}(3)$ rotation's matrix's first two rows. \\
\textbf{Action} We use delta end effector pose as the action parameterization. At each prediction step, the model predicts $t$ actions. Given \textit{absolute} end effector action transforms in $T_1, T_2, \cdots, T_t$ in a trajectory and the current end-effector pose $T_{\text{ee}}$, we define the relative transforms that the model needs to predict as $T_{\text{ee}}^{-1} T_1, T_{\text{ee}}^{-1} T_2, \cdots T_{\text{ee}}^{-1} T_t$. We then append the continuous absolute gripper position to each delta action. Similar to proprioception, we present the delta action by the relative end effector translation and a 6DoF rotation. This results in a 10-dimensional action representation. When rolling out the predicted actions, in addition to temporal ensembling~\cite{zhao2023learning}, we also use receding horizon control~\cite{diffusion}, and select an action horizon of 10 steps.

\subsection{System Information}
\reb{All models are trained on 4 NVIDIA A100 80GB GPUs. \algabbr pre-training on DROID takes 56 minutes and fine-tuneing on ICRT-MT takes 18 hours. ICRT-Llama7B takes roughly 28 hours to finetune.} We report the inference speed of \algabbr and \algabbr-Llama2 in \cref{tbl:inf_freq} averaged over 100 steps. All tests are performed on a workstation with NVIDIA RTX 3090Ti and Intel i5-12400F with 64GB memory. We find that using the proposed formulation, which can leverage the KV cache, we can run \algabbr-Llama2 at 10Hz naively. 
\tabInferenceFreq{h!}